\documentclass[pdftex,twocolumn,10pt,letterpaper]{extarticle}

\newcommand{\name}{3LC\xspace}
\newcommand{\AUTHORS}{Hyeontaek Lim, David G. Andersen, Michael Kaminsky}
\newcommand{\TITLE}{\name: Lightweight and Effective Traffic Compression for Distributed Machine Learning}
\newcommand{\KEYWORDS}{}
\newcommand{\CONFERENCE}{}
\newcommand{\PAGENUMBERS}{yes}       %
\newcommand{\COLOR}{yes}
\newcommand{\showComments}{yes}
\newcommand{\comment}[1]{}
\newcommand{\onlyAbstract}{no}

\usepackage[T1]{fontenc}
\usepackage[utf8]{inputenc}
\usepackage{newtxtext,newtxmath}       %
\usepackage{bm}                        %
\usepackage[scaled=0.92]{helvet}
\usepackage{inconsolata}              %

\usepackage{ifthen}
\ifthenelse{\equal{\PAGENUMBERS}{yes}}{%
\usepackage[nohead,
            left=1in,right=1in,top=1in,
            footskip=0.5in,bottom=1in,     %
            columnsep=0.25in
            ]{geometry}
}{%
\usepackage[noheadfoot,left=1in,right=1in,top=1in,
            footskip=0.5in,bottom=1in,
            columnsep=0.25in
	    ]{geometry}
}

\usepackage[font=bf]{caption}
\usepackage[compact]{titlesec}              %

\usepackage{enumitem}
\setlist{itemsep=0pt,parsep=0pt}             %

\usepackage{fancyhdr}

\ifthenelse{\equal{\PAGENUMBERS}{yes}}{%
  \pagestyle{plain}
}{%
  \pagestyle{empty}
}

\usepackage[numbers,sort]{natbib}

\usepackage{booktabs}
\usepackage{color}
\usepackage{colortbl}
\usepackage{float}                           %

\usepackage{subcaption} %

\ifthenelse{\equal{\COLOR}{yes}}{%
  \usepackage[colorlinks,citecolor=blue]{hyperref}%
}{%
  \usepackage[pdfborder={0 0 0}]{hyperref}%
}
\usepackage{url}
\makeatletter
\g@addto@macro{\UrlBreaks}{\UrlOrds}
\g@addto@macro{\UrlBreaks}{\do\/\do\a\do\b\do\c\do\d\do\e\do\f\do\g\do\h\do\i\do\j\do\k\do\l\do\m\do\n\do\o\do\p\do\q\do\r\do\s\do\t\do\u\do\v\do\w\do\x\do\y\do\z\do\A\do\B\do\C\do\D\do\E\do\F\do\G\do\H\do\I\do\J\do\K\do\L\do\M\do\N\do\O\do\P\do\Q\do\R\do\S\do\T\do\U\do\V\do\W\do\X\do\Y\do\Z\do\1\do\2\do\3\do\4\do\5\do\6\do\7\do\8\do\9\do\0\do\.}
\makeatother

\hypersetup{%
pdfauthor = {\AUTHORS},
pdftitle = {\TITLE},
pdfsubject = {\CONFERENCE},
pdfkeywords = {\KEYWORDS},
bookmarksopen = {true}
}

\pdftrailerid{} %
\definecolor{placeholderbg}{rgb}{0.85,0.85,0.85}

\usepackage[pdftex]{graphicx}
\usepackage{soul}
\soulregister\cite7
\soulregister\ref7
\soulregister\pageref7

\usepackage{xspace}
\usepackage{siunitx}

\usepackage[plain]{algorithm}
\usepackage{listings}

\definecolor{mygreen}{rgb}{0,0.6,0}
\definecolor{mygray}{rgb}{0.5,0.5,0.5}
\definecolor{mymauve}{rgb}{0.58,0,0.82}

\usepackage[absolute]{textpos}

\newfloat{acmcr}{b}{acmcr}

\newcommand{\note}[2]{%
    \ifthenelse{\equal{\showComments}{yes}}{\textcolor{#1}{#2}}{}%
}

\newcommand{\para}[1]{\noindent\textbf{#1}}

\date{}
\title{\textbf{\TITLE}}
\author{Hyeontaek Lim,$^1$ David G. Andersen,$^1$ Michael Kaminsky$^2$\\
\textit{$^1$Carnegie Mellon University, $^2$Intel Labs}}

\usepackage{silence}
\usepackage{balance}

\usepackage{microtype}  %

\begin{document}

\maketitle

\ifthenelse{\equal{\PAGENUMBERS}{yes}}{%
}{%
  \thispagestyle{empty}
}

\makeatletter{}%

\begin{abstract}

The performance and efficiency of distributed machine learning (ML) depends
significantly on how long it takes for nodes to exchange state changes.
Overly-aggressive attempts to reduce communication often sacrifice final model
accuracy and necessitate additional ML techniques to compensate for this loss,
limiting their generality.  Some attempts to reduce communication incur high
computation overhead, which makes their performance benefits visible only over
slow networks.

We present \emph{\name}, a lossy compression scheme for state change traffic
that strikes balance between multiple goals: traffic reduction, accuracy,
computation overhead, and generality.  It combines three new
techniques---\emph{3-value quantization with sparsity multiplication},
\emph{quartic encoding}, and \emph{zero-run encoding}---to leverage strengths of
quantization and sparsification techniques and avoid their drawbacks. It
achieves a data compression ratio of up to $39$--$107\times$, almost the same
test accuracy of trained models, and high compression speed.  Distributed ML
frameworks can employ \name without modifications to existing ML algorithms.
Our experiments show that \name reduces wall-clock training time of
ResNet-110--based image classifiers for CIFAR-10 on a 10-GPU cluster by up to
$16$--$23\times$ compared to TensorFlow's baseline design.

\end{abstract}

\ifthenelse{\equal{\onlyAbstract}{no}}{%
\makeatletter{}%

\section{Introduction}
\label{sec:intro}

Distributed machine learning (ML) harnesses high aggregate computational power
of multiple worker nodes.  The workers train an ML model by performing local
computation and transmitting \emph{state changes} to incorporate progress made
by the local computation, which are repeated at each
\emph{training step}.  Common metrics of interest in distributed ML include
\emph{accuracy} (how well a trained model performs) and \emph{training time}
(wall-clock time until a model reaches a trained state).  To improve training
time, distributed ML must be able to transmit large state change data quickly and
avoid impeding local computation.

However, the network does not always provide sufficient bandwidth for rapid
transmission of state changes.  Large-scale deployment of distributed ML often
require the workers to communicate over a low-bandwidth wide-area network (WAN)
to conform to local laws that regulate transferring sensitive training data
(e.g., personal photos) across regulatory
borders~\cite{Vulimiri:nsdi2015,Cano:arxiv2016,Hsieh:nsdi2017,www-eu-privacy-shield,www-china-cybersecurity-law}.
Some data might be pinned to mobile
devices~\cite{Konevcny:arxiv2016,McMahan:aistats2017}, forcing distributed ML to
use a slow and sometimes metered wireless network.  Recent performance studies
show that in-datacenter distributed training can demand more
bandwidth than local networks and even GPU interconnects currently
offer~\cite{Zhang:atc2017,Wen:nips2017,Alistarh:nips2017,Lin:arxiv2017}.

\emph{Communication reduction} intends to mitigate the network bottleneck by
reducing the overall communication cost.  In particular, lossy compression
schemes reduce the volume of state change data by prioritizing transmission of
important state
changes~\cite{Recht:nips2011,Ho:nips2013,Li:osdi2014,Abadi:osdi2016,Hsieh:nsdi2017}.
Unfortunately, existing schemes suffer one or more problems: They offer only a
small amount of network traffic reduction, sacrifice the accuracy of the trained
model, incur high computation overhead, and/or require modifications to
existing ML algorithms.

We present \name\footnote{Read as ``elk.''}~(3-value lossy compression), a
lightweight and efficient communication reduction scheme.  \name strikes a
balance between traffic reduction, accuracy, computation overhead, and
generality, to provide a ``go-to'' solution for bandwidth-constrained
distributed ML\@.  Our design (1)~uses only $0.3$--$0.8$~bits for each
real-number state change on average (i.e., traffic reduction by $39$--$107\times$ from
original 32-bit floating point numbers), (2)~causes small or no loss in accuracy
when using the same number of training steps, (3)~adds low computation overhead,
and (4)~runs with unmodified ML algorithms.

To achieve both high efficiency and high quality for distributed ML,
\name unifies two well-known lossy compression approaches commonly used for
communication reduction: \emph{Quantization} encodes state changes in low
resolution, and \emph{sparsification} only picks likely important parts of state
changes.  We do not blindly combine two approaches because doing so might end up
suffering drawbacks of both approaches; instead, we take their principle and
reconstruct them as a lightweight-yet-effective lossy compression scheme.

\name combines three new techniques:

\textbf{3-value quantization with sparsity multiplication} is a lossy
transformation that maps each floating-point number representing a state change
onto three values $\{-1, 0, 1\}$, with a knob that controls the compression level.
It corrects resulting quantization errors over time by using error
accumulation buffers.  Since it makes a small impact on the trained model's
accuracy, it does not require compensating for potential accuracy loss with
ML algorithm changes.

\textbf{Quartic encoding} is a lossless transformation that folds each group of
five 3-values into a single byte using fast vectorizable operations, which takes
20\% less space use than simple 2-bit encoding of 3-value data.  The quartic
encoding output is easy to compress further.

\textbf{Zero-run encoding} is a lossless transformation that shortens
consecutive runs of common bytes (groups of five zero values) by using a variant
of run-length encoding~\cite{Robinson1967} specialized for quartic encoded data.  It achieves
approximately a 2$\times$ or higher compression ratio, which varies by the distribution of
state change values.

Our empirical evaluation of \name and prior communication reduction techniques
on our custom 10-GPU cluster shows that \name is more effective in saving
traffic reduction while preserving high accuracy at low computation overhead.  When training image
classifiers based on ResNet-110~\cite{He2015} for the CIFAR-10
dataset~\cite{Krizhevsky:tr2009}, \name reduces training time to reach similar
test accuracy by up to $16$--$23\times$.  To measure \name's practical performance
gains over a strong baseline, we use a production-level distributed training
implementation on TensorFlow~\cite{Abadi:osdi2016} that is already optimized for
efficient state change transmission.

\makeatletter{}%
\section{Distributed ML Background}
\label{sec:background}

\begin{figure}
  \centering
  \includegraphics[width=0.95\columnwidth]{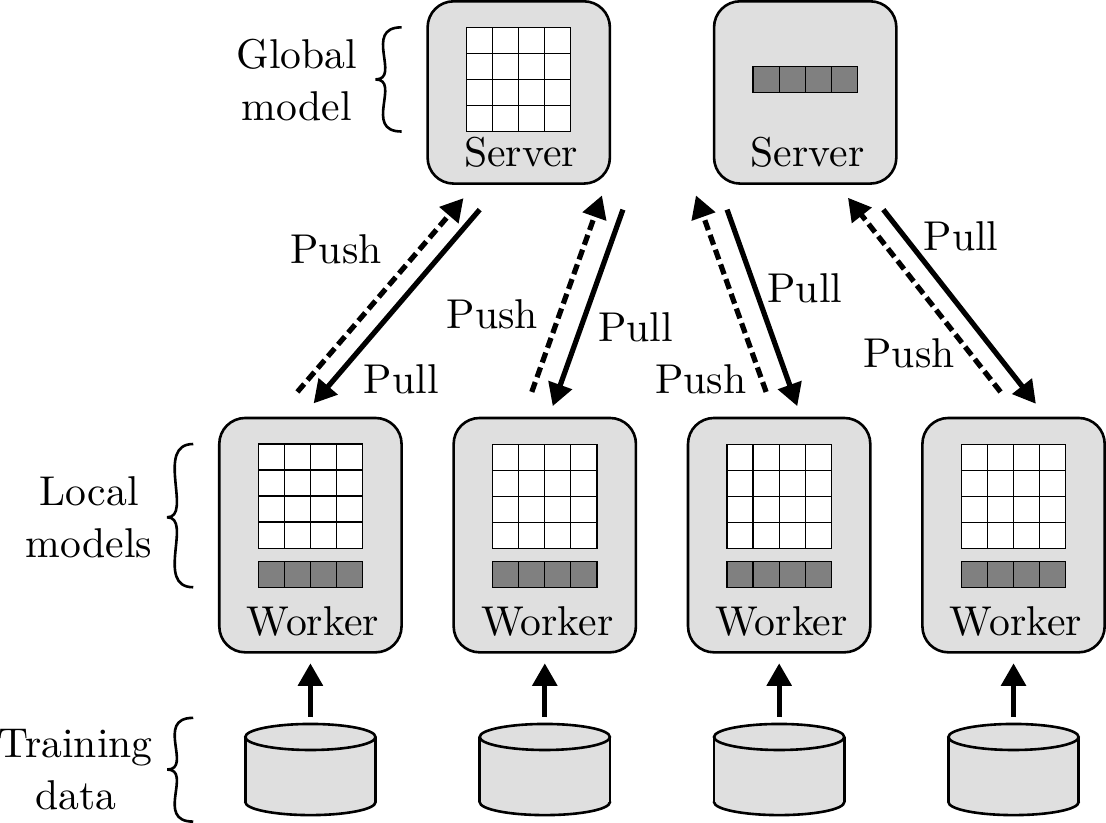}
  \caption{Distributed machine learning architecture using parameter servers.}
  \label{fig:dist-learning}
\end{figure}

Machine learning (ML) is a resource-heavy data processing task.  Training a
large-scale deep neural network (DNN) model may require tens of thousands of
machine-hours~\cite{Chilimbi:osdi2014}.  Distributed ML reduces
the total training time by parallelization~\cite{Li:osdi2014,Abadi:osdi2016}.

Figure~\ref{fig:dist-learning} depicts typical distributed DNN training using
parameter
servers~\cite{Ho:nips2013,Li:osdi2014,Chilimbi:osdi2014,Cui:eurosys2016}.
\emph{Parameter servers}, or simply \emph{servers}, store a partition of the
global model, which consists of \emph{parameters} (trainable variables).
\emph{Workers} keep a local copy of the model and training dataset.  The
parameters (and their state changes) are often represented as \emph{tensors}
(multidimensional arrays).  For example, the ``weights'' of a fully-connected
layer (a matrix multiply) would be a single 2-D tensor of floats.  The weights
of a different layer would be a separate 2-D tensor.

The workers train the model by repeatedly performing local computation and state
change transmission via the servers.  Each \emph{training step} includes the
following sub-steps: \emph{Forward pass:} The workers evaluate a \emph{loss
function} (objective function) for the current model using the local training dataset.
\emph{Backward pass:} The workers generate \emph{gradients} that indicate how
the model should be updated to minimize the loss function.  \emph{Gradient
  push:} The workers send the gradients to the servers.  \emph{Gradient
aggregation and model update:}  The servers average the gradients from the
workers and update the global model based on the aggregated gradients.
\emph{Model pull:} The workers retrieve from the servers \emph{model deltas}
that record the model changes, and apply the deltas to the local model.

Distributed ML may observe two types of communication costs, training step
barriers and state change traffic, which we discuss in the rest of this section.

\begin{figure*}[t]
  \centering
  \begin{subfigure}[b]{0.5\textwidth}
    \centering
    \includegraphics[scale=0.7]{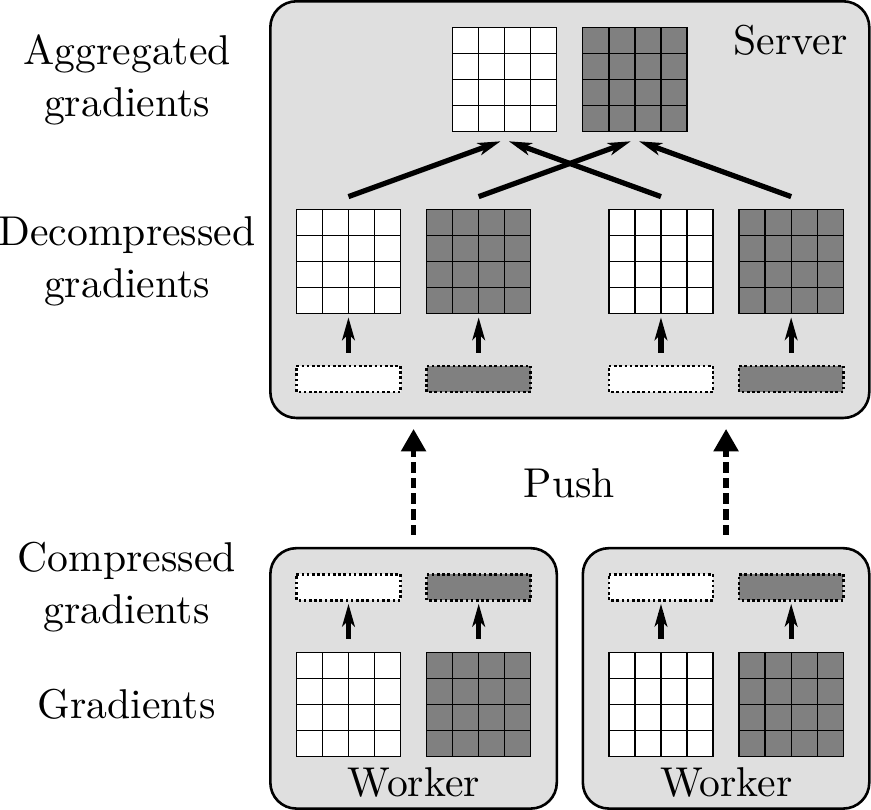}
    \caption{Gradient pushes from workers to servers.}
    \label{fig:push}
  \end{subfigure}%
  ~
  \begin{subfigure}[b]{0.5\textwidth}
    \centering
    \includegraphics[scale=0.7]{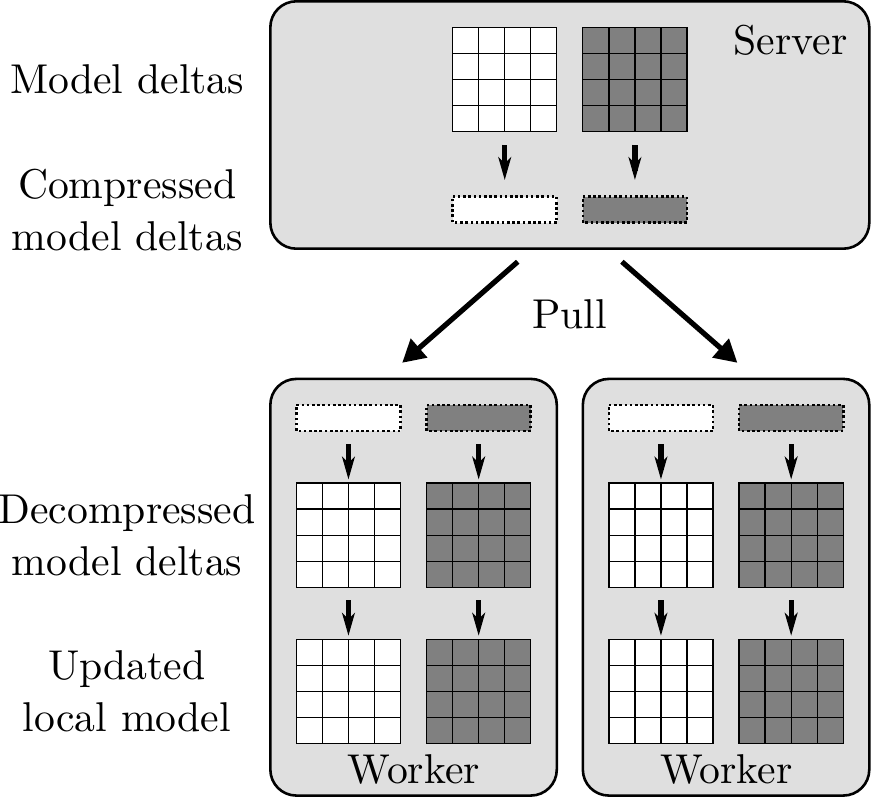}
    \caption{Model pulls from servers to workers.}
    \label{fig:pull}
  \end{subfigure}%
  \caption{Point-to-point tensor compression for two example layers in \name.}
  \label{fig:push-and-pull}
\end{figure*}

\subsection{Relaxing Barriers}
\label{sec:relaxed-barriers}

One important pillar of distributed ML research is how to perform efficient
synchronization of workers using barriers.  Although relaxing barriers is not the
main focus of our work, we briefly describe related techniques because modern
distributed ML systems already employ these optimizations to partially hide
communication latency.

In vanilla \emph{bulk synchronous parallel} (BSP), workers train on
an identical copy of the model~\cite{Valiant:cacm1990}.  BSP forces the servers
to wait for all workers to push gradients, and the workers to wait for the
servers to finish updating the global model before model pulls.  In this model,
slow or failed workers (``straggler'')~\cite{Recht:nips2011,Ho:nips2013} make
other workers waste computation resources, increasing training time.

To reduce a straggler problem, researchers have capitalized upon the property
that stochastic gradient descent and its variants commonly used in distributed
ML tolerate a small amount of inconsistency in the model across the
workers~\cite{Recht:nips2011}.  Fully asynchronous state change transmission
permits a worker to submit an update based on an arbitrarily stale version of the
model~\cite{Recht:nips2011}.  Approaches such as stale synchronous parallel
make a compromise between two extremes by limiting the maximum asynchrony
of the model for which an update is calculated~\cite{Ho:nips2013}.

A common downside of asynchronous state change transmission is that it may
accomplish less useful work per training step because of desynchronized local
models.  Asynchronous state change transmission generally requires more training
steps than BSP to train a model to similar test
accuracy~\cite{Recht:nips2011,Ho:nips2013,Li:osdi2014,Abadi:osdi2016,Hsieh:nsdi2017}.
Thus, recent distributed ML frameworks often advocate synchronous state change
transmission while using other techniques that mitigate
stragglers.  For instance,
TensorFlow~\cite{Abadi:osdi2016}'s stock distributed training implementation,
\texttt{SyncReplicasOptimizer}, uses backup workers: A global training step can
advance if a sufficient number of updates to the latest model have been
generated regardless of the number of unique workers that calculated the
updates~\cite{Chen:iclr2016}.

Modern distributed ML frameworks split barriers into more \emph{fine-grained
barriers} that help hide communication latency.  For example, Poseidon pushes
individual layers' gradients, allowing the servers to update part of the model
and let the workers pull that part instead of having to wait for the entire
model to be updated~\cite{Zhang:atc2017}.  TensorFlow's
\texttt{SyncReplicasOptimizer} pulls updated model data for individual layers as
they are evaluated in the forward pass.  Such fine-grained barriers facilitate
overlapping communication and computation and improve computational efficiency
of distributed ML\@.

\subsection{Compressing State Change Traffic}

Relaxed barriers reduce communication costs, but they do not completely hide
communication latency.  Gradient pushes and model pulls are sensitive to the
available network bandwidth, as these steps need to transmit large data quickly,
and state change transmission can take longer as the model size grows and/or the
network bandwidth is more
constrained~\cite{Zhang:atc2017,Wen:nips2017,Alistarh:nips2017,Lin:arxiv2017,Hsieh:nsdi2017}.
If the transmission takes excessive time, cluster nodes experience long stall
time, harming the efficiency of distributed learning.

Quantization and sparsification techniques make state change transmission
generate less network traffic by applying lossy compression to the state change
data.  They prioritize sending a small amount of likely important state change
information and defer sending or even ignore unsent changes.
\emph{Quantization} uses low-resolution values to transmit the approximate
magnitude of the state change
data~\cite{Seide:interspeech2014,Alistarh:nips2017,Wen:nips2017}.
\emph{Sparsification} discovers state changes with large magnitude and transmits
a sparse version of tensors that contain these state
changes~\cite{Li:osdi2014,Wei:socc2015,Watcharapichat:socc2016,Hsieh:nsdi2017,Aji:emnlp2017,Lin:arxiv2017}.

Note that quantization and sparsification we discuss in this paper differ from
model compression~\cite{Han:iclr2016,Jouppi:isca2017}.  \emph{Model compression}
reduces the memory requirement and computation cost of DNN models by quantizing
and reducing their parameters (not state changes).  \emph{Inference} with a
compressed model can run faster without demanding much computation and memory
resources.  In contrast, our paper focuses on \emph{distributed training} of a
model that consists of full-precision parameters, which can be processed using
model compression after training finishes.

\makeatletter{}%

\section{Design}
\label{sec:design}

The design goal of \name is to achieve good balance between traffic reduction,
accuracy, computation overhead, and generality.  We present the high-level
design of \name and its components in detail.

\name is a point-to-point tensor compression scheme.
Figure~\ref{fig:push-and-pull} depicts how \name compresses, transmits, and
decompresses state change tensors for two example layers.  One compression context
encompasses the state for compression and decompression of a single tensor that
represents gradients (a push from a worker to a server) or model deltas (a pull
from a server to a worker) of a single layer in a deep neural network.

This point-to-point design preserves the communication pattern of existing
parameter server architectures.  It adds no extra communication channels between
servers or workers because it involves no additional coordination between
them.  Some designs~\cite{Wen:nips2017} synchronize their
compression parameters among workers before actual traffic compression, which
adds round trips to communication between the workers for each training step.

A potential performance issue of this point-to-point compression is redundant
work during model pulls. Servers send identical data to workers so that the workers
update their local model to the same state.  If the servers compress individual
pulls separately, it would perform redundant compression work.  \name optimizes
model pulls by sharing compression: The servers compresses model deltas and make a
shared local copy of the compressed model deltas, and the workers pull the
compressed data as if they pull uncompressed model deltas
(Figure~\ref{fig:pull}).  Note that distributed ML frameworks that allow loosely
synchronized local models on
workers~\cite{Recht:nips2011,Ho:nips2013,Li:osdi2014,Hsieh:nsdi2017} may require
multiple copies of compressed model deltas, each of which is shared by a subset
of the workers with the same local model.

For \name's tensor compression and decompression, we introduce one lossy and two
lossless transformations: 3-value quantization with sparsity multiplication
(Section~\ref{sec:design-3vq}), quartic encoding (Section~\ref{sec:design-qe}),
and zero-run encoding (Section~\ref{sec:design-zre}).  The rest of this section
describes their designs and rationale.

\begin{figure}[t]
  \centering
  \includegraphics[width=0.95\columnwidth]{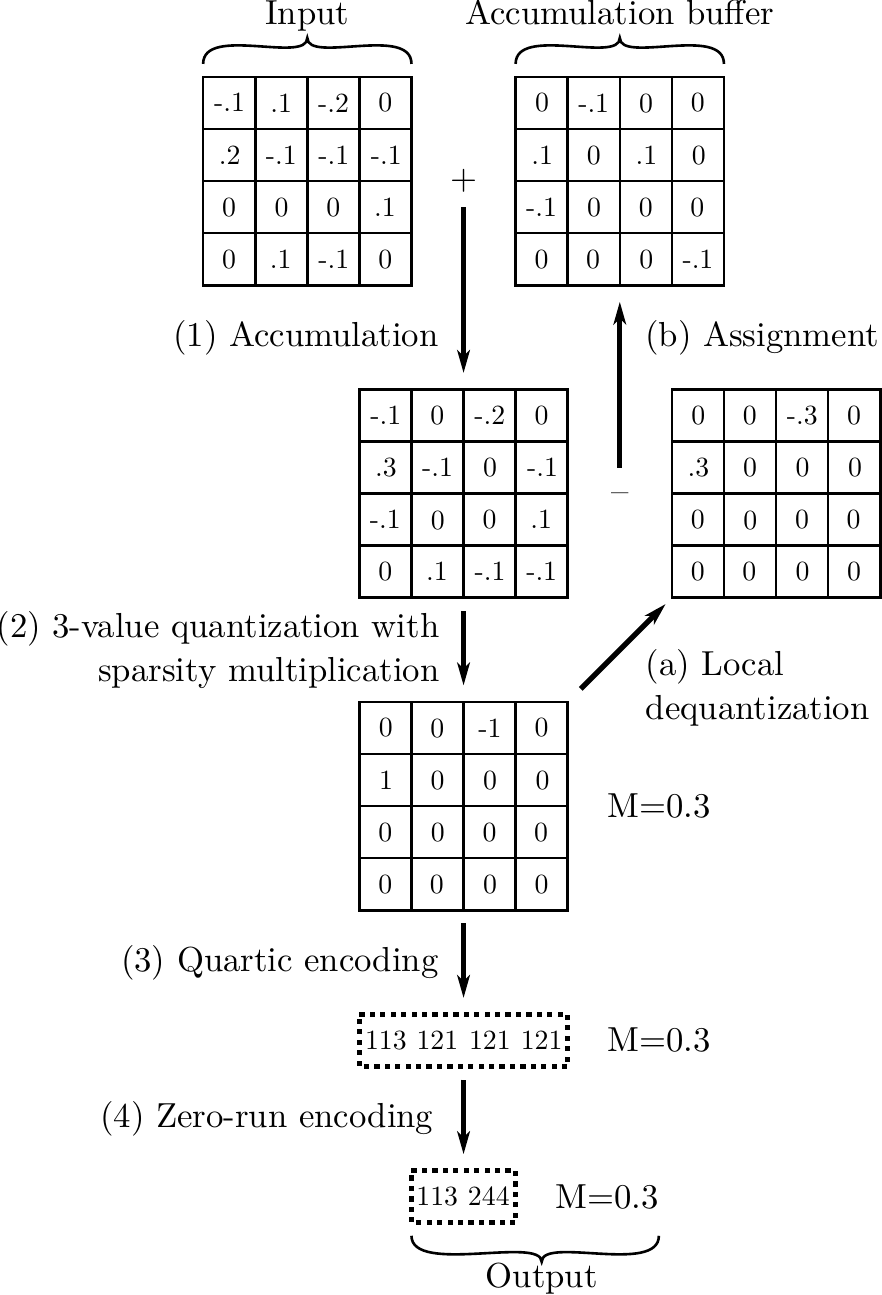}
  \caption{Tensor compression in \name.}
  \label{fig:compression}
\end{figure}

\subsection{3-value Quantization with\\Sparsity Multiplication}
\label{sec:design-3vq}

\emph{3-value quantization} compresses a state change tensor by leveraging the
distribution of state changes that are centered around zero~\cite{Wen:nips2017}.  It
transforms a full-precision input tensor into a new tensor of three discrete
values $\{-1, 0, 1\}$ that has the same shape (dimensions) as the input tensor,
and a full-precision scalar $M$ that is the maximum magnitude of the input
tensor values scaled by a \emph{sparsity multiplier} $s$ ($1 \le s < 2$).

Suppose $T_\textrm{in}$ is an input tensor.  The output of 3-value
quantization is
\begin{eqnarray}
  M & = & \textrm{max}(|T_\textrm{in}|) \cdot s \label{eqn:M} \\
  T_\textrm{quantized} & = & \textrm{round}\left(\frac{T_\textrm{in}}{M}\right) \label{eqn:Q}
\end{eqnarray}

Dequantization is a simple multiplication:
\begin{eqnarray}
  T_\textrm{out} & = & M \cdot T_\textrm{quantized} \label{eqn:DQ}
\end{eqnarray}

$s$ controls the compression level of \name.  $s=1$ is the default multiplier
that preserves the maximum magnitude of values in the input tensor across
quantization and dequantization.  With a larger $s$ ($1 < s < 2$), the
quantization output is sparser (more zeros) because the magnitude of more values
are smaller than $M/2$.  The sparser output may contain less state change
information, but can be compressed more aggressively by zero-run encoding.

Quantization followed by dequantization returns a slightly different tensor from
the input tensor, causing \emph{quantization errors}.  \name can experience
relatively larger quantization errors especially when $s$ is larger because
dequantization can make a value farther from its original value (but within a
certain limit to ensure convergence).

\name corrects quantization errors using error accumulation
buffers~\cite{Seide:interspeech2014,Wei:socc2015,Watcharapichat:socc2016,Hsieh:nsdi2017,Aji:emnlp2017}.
It allows quantization errors to occur in the first place, but attempts to
correct in quantization at later training steps.  It keeps a local per-tensor error
accumulation buffer to remember the errors across training steps.

Figure~\ref{fig:compression} depicts 3-value quantization with error
accumulation, using $s=1$.  Step~(1) accumulates the input tensor into a local
buffer.  Step~(2) applies 3-value quantization to the sum.  Step~(a) dequantizes the
quantized data locally.  Step~(b) calculates remaining quantization errors and
stores them in the local buffer.

\para{Alternative quantization techniques:} \emph{Stochastic quantization}
outputs randomized quantization values whose expectation matches their input
value~\cite{Alistarh:nips2017}. It eliminates biases that exist in deterministic
rounding.  For instance, TernGrad~\cite{Wen:nips2017} uses three values for
quantization similarly to 3-value quantization (without the sparsity
multiplication), but uses stochastic selection of output values.  We decided to
use error accumulation buffers instead of stochastic quantization for several
reasons: (1)~Biases that are caused by non-stochastic quantization can be
corrected over time by using error accumulation buffers.  (2)~When used alone,
error correction with error accumulation buffers achieves better accuracy than
stochastic quantization in our evaluation (Section~\ref{sec:eval}); designs
using stochastic quantization require more bits for
quantization~\cite{Alistarh:nips2017} or additional accuracy-compensation
techniques~\cite{Wen:nips2017} for high accuracy.  (3)~Using both error
accumulation buffers and stochastic quantization caused training fail to converge in our
experiments.

\emph{Squared quantization error minimization} is a deterministic method that
picks magnitude values that minimize the squared sum of quantization errors.
For instance, 1-bit stochastic gradient descent maps non-negative values and
negative values of an input tensor into two values $\{0, 1\}$, and each of these
two values are dequantized using a different $M$ value that is the average of
non-negative or negative values in the input
tensor~\cite{Seide:interspeech2014}.  In designing \name, we avoid reducing the
magnitude of quantized values instead of pursuing minimum squared quantization
errors because (1)~low quantization errors do not necessarily lead to high
accuracy in empirical evaluation (Section~\ref{sec:eval}) and (2)~other lossy
compression techniques for state change traffic also preserve the approximate
magnitude of input tensors for better accuracy even though doing so may provide
weaker theoretic guarantees~\cite{Wen:nips2017,Alistarh:nips2017}.

\para{Alternative sparsification techniques:} The sparsity multiplier plays a
role similar to the \emph{threshold} knob in sparsification-based compression
techniques~\cite{Hsieh:nsdi2017,Lin:arxiv2017}. Both affect how many distinct
state changes are chosen for transmission.  However, thresholding makes a
decompressed tensor have much smaller average values than the input tensor by
omitting many input values (even though they are small);
overly-aggressive thresholding can result in lower accuracy, and compensating
for it requires changing ML algorithms such as modified momentum
calculation~\cite{Lin:arxiv2017} that does not generalize well to non-gradient
data transmission such as model pulls.  In contrast, dequantization using
sparsity multiplication enlarges (now scarcer) large values, better preserving
the average magnitude of the input tensor.

3-value quantization always uses a dense form (array) of tensors. Dense tensor
operations are easier to accelerate than sparsification-based compression
techniques that requires dense-to-sparse and sparse-to-dense tensor conversion
whose vectorization is often unavailable (e.g., TensorFlow~\cite{Abadi:osdi2016}
has only a non-vectorized CPU implementation and no GPU implementation for
sparse-to-dense conversion as of February 2018).

Prior lossy traffic reduction schemes often employ custom rounding
function~\cite{Li:osdi2014,Seide:interspeech2014,Aji:emnlp2017,Lin:arxiv2017,Hsieh:nsdi2017}
that often makes vectorization difficult.  3-value quantization instead uses
simple $\textrm{round}()$ whose vectorized version is readily available on
modern CPUs and GPUs~\cite{www-intel-dev,www-nvidia-cuda-c-programming-guide}.

\para{Convergence:} 3-value quantization with sparsity multiplication retains
convergence of state change tensors.  $\textrm{round}()$ adds a maximum absolute error
of $1/2$.  By Equations~\ref{eqn:Q} and~\ref{eqn:DQ}, the maximum absolute error
$max(|T_\textrm{in}-T_\textrm{out}|)$ is bounded by $M/2$.  Note $M/2 <
\textrm{max}(|T_\textrm{in}|)$ because of Equation~\ref{eqn:M} and $1 \le s <
2$.  Let $\alpha$ be a decaying learning rate (if $T_\textrm{in}$ is a gradient
tensor) or $1$ (if $T_\textrm{in}$ is a model delta tensor).  Under an
assumption that $\alpha T_\textrm{in}$ converges to zero, $\alpha M/2$ converges
to zero, and $\alpha T_\textrm{out}$ also converges to zero.

\subsection{Quartic Encoding}
\label{sec:design-qe}

Compactly encoding 3-values is nontrivial because CPU and GPU architectures do
not provide native data types for base-3 numbers.  The space requirement of a
simple encoding for 3 discrete values using 2~bits~\cite{Wen:nips2017} is larger
than the theoretic minimum of $\log_2{3}\approx1.585$ by approximately $26$\%.

Quartic encoding is a fixed-length representation for a 3-value quantized
tensor.  It takes five 3-values and packs them into a single byte
[Figure~\ref{fig:compression} Step~(3)], using \emph{1.6~bits per 3-value} that
is only $0.95$\% higher than the theoretic bound.  Quartic encoding exploits the
fact that a quartic-form expression $a \cdot 3^4 + b \cdot 3^3 + c \cdot 3^2 + d
\cdot 3^1 + e$ has only $3^5=243$ distinct values ($\le 256$) if $a, \dots, e
\in \{0, 1, 2\}$.  Quartic encoding of a 3-value quantized tensor takes the
following steps:
\begin{enumerate}
  \item Element-wise add 1 to the 3-value quantized tensor
  \item Type cast it to an unsigned 8-bit integer array
  \item Flatten it into a 1-D array
  \item Pad it with zeros to make its length a multiple of 5
  \item Divide the array into 5 partitions: $p_0, p_1, p_2, p_3, p_4$
  \item Compute $a = p_0 \cdot 81 + p_1 \cdot 27 + p_2 \cdot 9 + p_3 \cdot 3 + p_4$
\end{enumerate}

Decoding reverses encoding steps:
\begin{enumerate}
  \item Restore $p_0, p_1, p_2, p_3, p_4$ by dividing $a$ by a power of 3 and taking the remainder (a base-3 conversion)
  \item Concatenate $p_0, p_1, p_2, p_3, p_4$
  \item Unpad, reshape, and type cast
  \item Element-wise subtract 1
\end{enumerate}

These encoding and decoding steps can be easily vectorized on CPUs and GPUs using
operations provided by ML frameworks.

\subsection{Zero-run Encoding}
\label{sec:design-zre}

The input to quartic encoding is sparse (even though the data structure is
dense), containing a large number of zeros. The number of zeros increases as the
sparsity multiplier $s$ increases.  Although quartic encoding is compact, it
always generates a fixed-length representation, which does not take advantage of
the sparseness in the input.

Zero-run encoding is a variant of run-length encoding~\cite{Robinson1967}, but
is specialized to quartic-encoded data.  Note that quartic encoding maps a group
of five zero values from the 3-value quantized tensor into a byte value
\texttt{121}.  Also recall that quartic encoding only outputs byte values of
\texttt{0}--\texttt{242}.  Zero-run encoding finds a run of \texttt{121} and
replaces it with a new byte value between \texttt{243} and \texttt{255},
inclusive [Figure~\ref{fig:compression} Step~(4)].  In other words, $k$
consecutive occurrences of \texttt{121} ($2 \le k \le 14$) are replaced with a
single byte value of \texttt{243+(k-2)}.  In a hypothetical case of compressing
a zero 32-bit floating-point tensor, the combination of all techniques in \name
reaches a compression ratio of $280\times$.

Compared to general-purpose compression algorithms or entropy coding
schemes~\cite{www-snappy,Oland:icassp2015,Alistarh:nips2017}, zero-run encoding
is simple to implement and fast to run by avoiding any bit-level operation and
lookup tables, which helps \name keep low computation overhead.

\makeatletter{}%
\section{Implementation}
\label{sec:impl}

We implement a prototype of \name on TensorFlow~\cite{Abadi:osdi2016}.  3-value quantization
with sparsity multiplication and quartic encoding use TensorFlow's built-in
vectorized operators. Zero-run encoding uses a custom operator written in C++.

Our prototype includes a distributed optimizer that retains the interface of
\texttt{SyncReplicasOptimizer}, which is TensorFlow's stock distributed training
implementation. The distributed optimizers augment any local optimizer with
distributed training by providing gradient aggregation and training step
barriers.  To replicate TensorFlow's tensor caching and incremental pull
behavior that copies each remote tensor into a local cache before local access
to that tensor, our prototype ensures that first-time access to a tensor at each
training step executes extra operators that pull, decompress, and apply model
deltas to the tensor.

One user-facing change is tensor allocation.  Our prototype asks the user
program to call a helper function that provides the same interface as
\texttt{get\_variable()}, which is a TensorFlow function that allocates a single
tensor.  This helper function reserves buffers for error accumulation and compressed
model deltas, and assigns a correct physical location to the buffers.  The user
program can keep using the default \texttt{get\_variable()} for the tensors that
do not require compression (e.g., tensors for small layers); our distributed
optimizer falls back to \texttt{SyncReplicasOptimizer}'s behavior for
distributed training of these tensors.

\section{Evaluation}
\label{sec:eval}

We experimentally evaluate \name to quantify its effectiveness
against other communication reduction schemes.  Our experiments investigate
the following aspects:
\begin{itemize}
  \item Traffic: How much traffic does each scheme save?
  \item Training time: How much wall-clock training time do they save?
  \item Accuracy: What is the highest test accuracy each scheme can achieve using standard training steps?
  \item Convergence speed: What is the highest test accuracy they achieve using much fewer training steps?\footnote{High convergence speed with few training steps can be useful for
      accurate and fast hyperparameter (training configuration) optimizations using small
    computational resources~\cite{Snoek:nips2012,Golovin:kdd2017}.}
  \item Computation overhead: How low is their computation overhead?
\end{itemize}

\subsection{Compared Designs}

Our evaluation compares representative communication reduction schemes that we
implement on TensorFlow:

\texttt{32-bit float} is the baseline that transmits 32-bit floating-point state
changes without compression.

\texttt{8-bit int} is an 8-bit quantization scheme that approximates Google
Tensor Processing Unit's internal 8-bit quantization~\cite{Jouppi:isca2017}.
Our implementation uses 255 distinct values ($[-127,127]$, leaving $-128$
unused).

\texttt{Stoch 3-value + QE} uses stochastic 3-value quantization similar to
TernGrad (but without ``gradient clipping'')~\cite{Wen:nips2017}, and our
quartic encoding for 1.6-bit quantization (smaller than TernGrad's 2-bit
quantization).

\texttt{MQE 1-bit int} performs 1-bit quantization with minimum squared quantization
errors and error feedback~\cite{Seide:interspeech2014}.

\texttt{25\% sparsification} and \texttt{5\% sparsification} choose 25\% and 5\%
of the largest state changes in each tensor, respectively, and accumulate unsent changes in buffers,
which reproduce common sparsification
techniques~\cite{Li:osdi2014,Wei:socc2015,Hsieh:nsdi2017,Aji:emnlp2017,Lin:arxiv2017}.
We use the magnitude (not relative magnitude~\cite{Hsieh:nsdi2017}) of values to
find largest values for better accuracy in our experiments.  To avoid exhaustive
sorting while finding a threshold, we only sort sampled input
values~\cite{Aji:emnlp2017}.  We use a bitmap to indicate which state
changes sparsification has selected, which adds 1~bit per state change as
traffic overhead regardless of input size.

\texttt{2 local steps} transmits state changes every 2 local steps.  Unsent
updates are accumulated locally and sent at the next training step using error
accumulation buffers.  It reduces the traffic almost by half and effectively
doubles the global batch size of distributed training.

\texttt{\name} is the full \name design. $s$ is the sparsity multiplier.

Similar to prior work~\cite{Alistarh:nips2017}, we exclude state changes for
small layers (batch normalization~\cite{Ioffe:icml2015} in our experiments) from
compression because avoiding computation overhead far outweighs compacting
already small tensors.

Note that the implementation of some compared designs are not identical to
prior proposed designs because their design is incompatible with our workload
and the TensorFlow parameter server architecture.  For instance,
\texttt{sparsification} does not use modified momentum
algorithms~\cite{Lin:arxiv2017} because and TensorFlow sends not only gradients,
but also model deltas to which their modifications of ML
algorithms are inapplicable.

\subsection{Evaluation Setup}

\para{Workload:} Our experiments train image classifiers based on
ResNet-110~\cite{He2015} for the CIFAR-10 dataset~\cite{Krizhevsky:tr2009}.
CIFAR-10 contains 50,000 training images and 10,000 testing images, each of which
has one of 10 labels.  ResNet-110 is a 110-layer convolutional neural network
for CIFAR-10.

\para{Detailed training configuration:} The following paragraphs provide an exhaustive
description of our ML parameters and environment for completeness.  We use
standard configurations and values from the
literature~\cite{He2015,Loshchilov:iclr2017}. Readers can feel free to skip
these ML-focused details and resume at the ``Hardware and Network'' paragraph.

We reuse the local optimizer type and hyperparameters for ResNet-110 training
from the original ResNet paper~\cite{He2015} except for the learning rate
schedule.  The local optimizer is TensorFlow's \texttt{MomentumOptimizer} with
the momentum of $0.9$.  The weight decay is $0.0001$.  We vary the learning rate
from 0.1 to 0.001, following the original learning rate range, but we use cosine
decay without restarts~\cite{Loshchilov:iclr2017} instead of the original
stepwise decay because the cosine decay achieves better
accuracy~\cite{Loshchilov:iclr2017,Bello:icml2017} and has fewer hyperparameters
to tune.  We apply the standard data augmentation that randomly crops and
horizontally flips original images to generate training examples~\cite{He2015}.

Our distributed training configuration follows the guideline for large-batch
stochastic training~\cite{Goyal:arxiv2017}.  We use a per-worker batch size of
32 images~\cite{Lin:arxiv2017}; using the original batch size of 128 reduces
accuracy for all designs because it produces a large global batch size of 1,280
on a 10-worker cluster.  We scale the learning rate proportionally to the worker
count and make one worker responsible for updating batch normalization
parameters~\cite{Goyal:arxiv2017}.  Our accuracy matches or exceeds the
accuracy of a ResNet-110 trained using a similar batch size but stepwise
decay~\cite{Lin:arxiv2017}.

We choose to train a ResNet because it is both a representative and
\emph{challenging} workload for communication reduction schemes to show their
performance benefits.  The ResNet architecture's ``identity mappings'' are
commonly found in high-accuracy neural network
architectures~\cite{Zoph:arxiv2017}.  Compared to traditional neural network
architectures such as VGG~\cite{simonyan2014very}, ResNet models typically have
small parameter count to computation ratios~\cite{Zoph:arxiv2017}, generating
\emph{less state change traffic} for the same amount of communication.  Its very
deep network structure permits \emph{efficient incremental transmission} of
state changes (Section~\ref{sec:relaxed-barriers}), facilitating overlapping
computation and communication and hiding communication latency.  Therefore, we
believe that a capability to show performance gains on the ResNet architecture
is likely to be transferable to other neural network architectures.

\para{Hardware and Network:} Our distributed training runs on a custom GPU
cluster.  It uses 10 workers with a GPU; each pair of workers shares a physical
machine equipped with two Intel Xeon E5-2680 v2 CPUs (total 20 physical cores),
128~GiB DRAM, and two Nvidia GTX 980 GPUs.  Our experiments use \texttt{numactl}
for CPU and memory isolation between worker pairs and
\texttt{CUDA\_VISIBLE\_DEVICES} for a dedicated per-worker GPU\@. A separate
machine acts as a parameter server.  We use the Linux Traffic
Control~\cite{www-linux-tc} on worker and server nodes to emulate constrained
network bandwidth.

\para{Measurement Methodology:} A dedicated node reads the snapshot of the
global model and calculates the \emph{top-1 score of the testing images} as test
accuracy.

Due to limited computation resources, we divide the experiments into two
categories of full measurement and accelerated measurement.  \emph{Full
measurement} measures training time, average per-step training time, and
accuracy on 1~Gbps by executing \emph{standard training steps} ($163.84$
epochs~\cite{He2015}, which is equivalent to 25,600 steps for 10~workers with a
batch size of 32).  \emph{Accelerated measurement} only obtains average per-step
time on 10~Mbps and 100~Mbps links by executing 100 and 1000 steps, respectively
(about 1~hour of training for \texttt{32-bit float}); one exception is that any
design with zero-run encoding runs 10\% of standard training steps to faithfully
reflect its compression ratios changing over time.
The learning rate schedule uses adjusted training steps as the total training
steps (as usual) to ensure each training run to sweep the entire learning rate
range.

\begin{table*}[t]
  \small
  \centering
  \begin{tabular}{l | S[table-format=3.2] S[table-format=3.2] S[table-format=3.2] | S[table-format=3.2] S[table-format=3.2,retain-explicit-plus]}
           & \multicolumn{3}{c|}{Speedup ($\times$)} & {} & {} \\
    Design                           & {@ 10 Mbps} & {@ 100 Mbps} & {@ 1 Gbps} & {Accuracy (\%)} & {Difference} \\ \hline\hline
    \texttt{32-bit float}            & 1           & 1            & 1          & 93.37           & {}     \\ \hline
    \texttt{8-bit int}               & 3.62        & 3.47         & 1.51       & 93.33           & -0.04  \\
    \texttt{Stoch 3-value + QE}      & 12.3        & 7.51         & 1.53       & 92.06           & -1.31  \\
    \texttt{MQE 1-bit int}           & 14.6        & 7.40         & 1.30       & 93.21           & -0.16  \\
    \texttt{25\% sparsification}     & 3.25        & 3.11         & 1.33       & 93.40           & +0.03  \\
    \texttt{5\% sparsification}      & 8.98        & 6.62         & 1.44       & 92.87           & -0.50  \\
    \texttt{2 local steps}           & 1.92        & 1.87         & 1.38       & 93.03           & -0.34  \\ \hline
    \texttt{\name (s=1.00)}          & 15.9        & 7.97         & 1.53       & 93.32           & -0.05  \\
    \texttt{\name (s=1.50)}          & 20.9        & 8.70         & 1.53       & 93.29           & -0.08  \\
    \texttt{\name (s=1.75)}          & 22.8        & 9.04         & 1.53       & 93.51           & +0.14  \\
    \texttt{\name (s=1.90)}          & 22.8        & 9.22         & 1.55       & 93.10           & -0.27  \\
  \end{tabular}
  \caption{Speedup over the baseline and test accuracy using standard training steps (graphs in the next page).}
  \label{tbl:tradeoff}
\end{table*}

\emph{We predict the training time on 10~Mbps and 100~Mbps by scaling the
training time from the 1~Gbps full measurement based on per-step training time
differences between full and accelerate measurement results while reusing the
accuracy from the full measurement.} Suppose a full measurement result for
1~Gbps is training time of $t_\textrm{full}$, per-step training time of
$s_\textrm{full}$, and an accelerated measurement result for 10~Mbps is per-step
training time of $s_\textrm{short}$.  We estimate the training time of 10~Mbps
to be $t_\textrm{full} \cdot s_\textrm{short} / s_\textrm{full}$.  We take test
accuracy obtained in the full measurement as-is because network bandwidth
changes do not affect test accuracy.  Without training time extrapolation,
obtaining a single datapoint on a slow network takes approximately 10 days on
our cluster, which would make it hard for us to compare many designs extensively
at high confidence.

We show the average of measurement results from multiple independent runs.  Each
experiment configuration is run 5 times for full measurement, and 3 times for
accelerated measurement.

\subsection{Macrobenchmark}

We examine the tradeoff between total training time and accuracy of compared
schemes.  Each datapoint on the graph represents a separate experiment
configuration; the learning schedule (the cosine decay) depends on total
training steps, requiring a new experiment for accuracy measurement for a
different number of total training steps.

Table~\ref{tbl:tradeoff} summarizes training time speedups over the
baseline and test accuracy when using standard training steps.  \name achieves
the best speedup across all network configurations, and its accuracy remains
similar to the baseline, except \texttt{\name (s=1.90)} that performs highly
aggressive traffic compression.  Other designs offer less training time
reduction or suffer lower accuracy.

Figure~\ref{fig:tradeoff-10} plots total training time and test accuracy on
10~Mbps when varying the total number of training steps to 25\%, 50\%, 75\%, and
100\% of standard training steps.  An experiment using 100\% training steps
gives the accuracy of fully trained models, while using fewer training steps
indicates the convergence speed of a design.

The result shows that designs that achieve high accuracy with many training
steps do not always yield high accuracy with fewer training steps.
\texttt{\name (s=1.75)} provides the best training time and maintains high
accuracy when using 100\% training steps because of its effective traffic
compression.  When using fewer training steps, \texttt{\name (s=1.00)} achieves
better accuracy.  \name's sparsity multiplication affects tradeoffs between
traffic reduction and convergence speed, but it does not necessarily harm
accuracy obtained using sufficient training steps (e.g., executing as many
training steps as standard no-compression training uses).

Note that \texttt{Stoch 3-value + QE} has lower accuracy than \name.  This
accuracy loss by stochastic quantization supports our design decision of using
error accumulation buffers to correct quantization errors.

With a faster network of 100~Mbps, as shown in Figure~\ref{fig:tradeoff-100}, the
benefit of reducing traffic begins to diminish and preserving high accuracy
becomes more important.  For example, \texttt{5\% sparsification} provides
always better speed-accuracy tradeoffs than \texttt{Stoch 3-value + QE},
which is different on 10~Mbps.

On a 1~Gbps network, in Figure~\ref{fig:tradeoff-1000}, the most influential
factors to speed-accuracy tradeoffs are high accuracy and low computation
overhead, and traffic reduction becomes less important.  \name provides high
accuracy using 75\% to 100\% of standard training steps and slightly lower
accuracy than \texttt{8-bit int} using fewer training steps.  \texttt{MQE 1-bit
int} is slower than \texttt{8-bit int} that transmits 8$\times$ more traffic;
the long training time of \texttt{MQE 1-bit int} is attributable to its high
computation overhead of using an unconventional rounding function.
\name does not add such high overhead because it leverages existing vectorized
operations.

\begin{figure*}[p]
  \centering
  \begin{subfigure}[b]{0.5\textwidth}
    \centering
    \includegraphics[width=\textwidth]{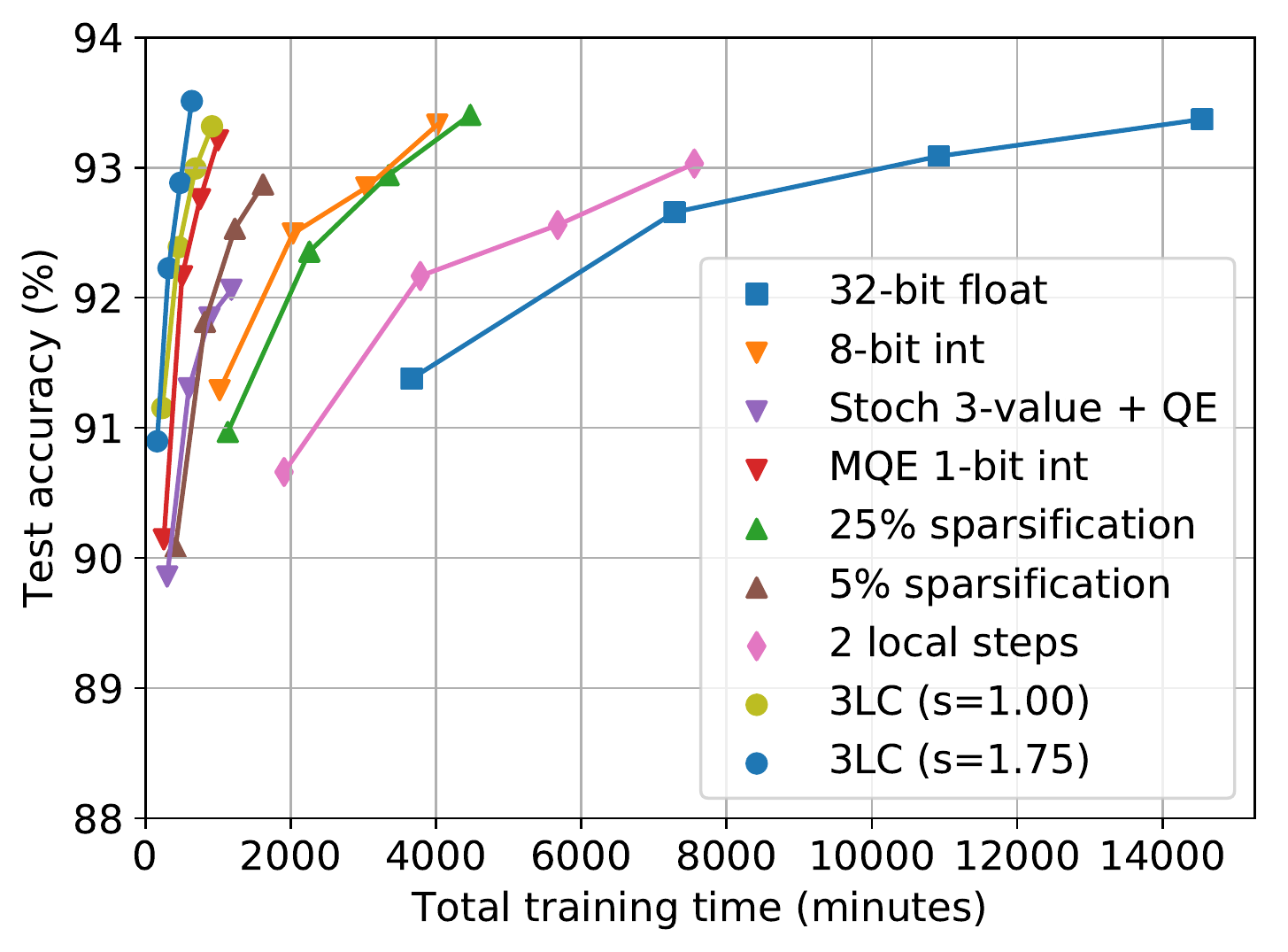}
    \caption{Overview}
  \end{subfigure}%
  ~
  \begin{subfigure}[b]{0.5\textwidth}
    \centering
    \includegraphics[width=\textwidth]{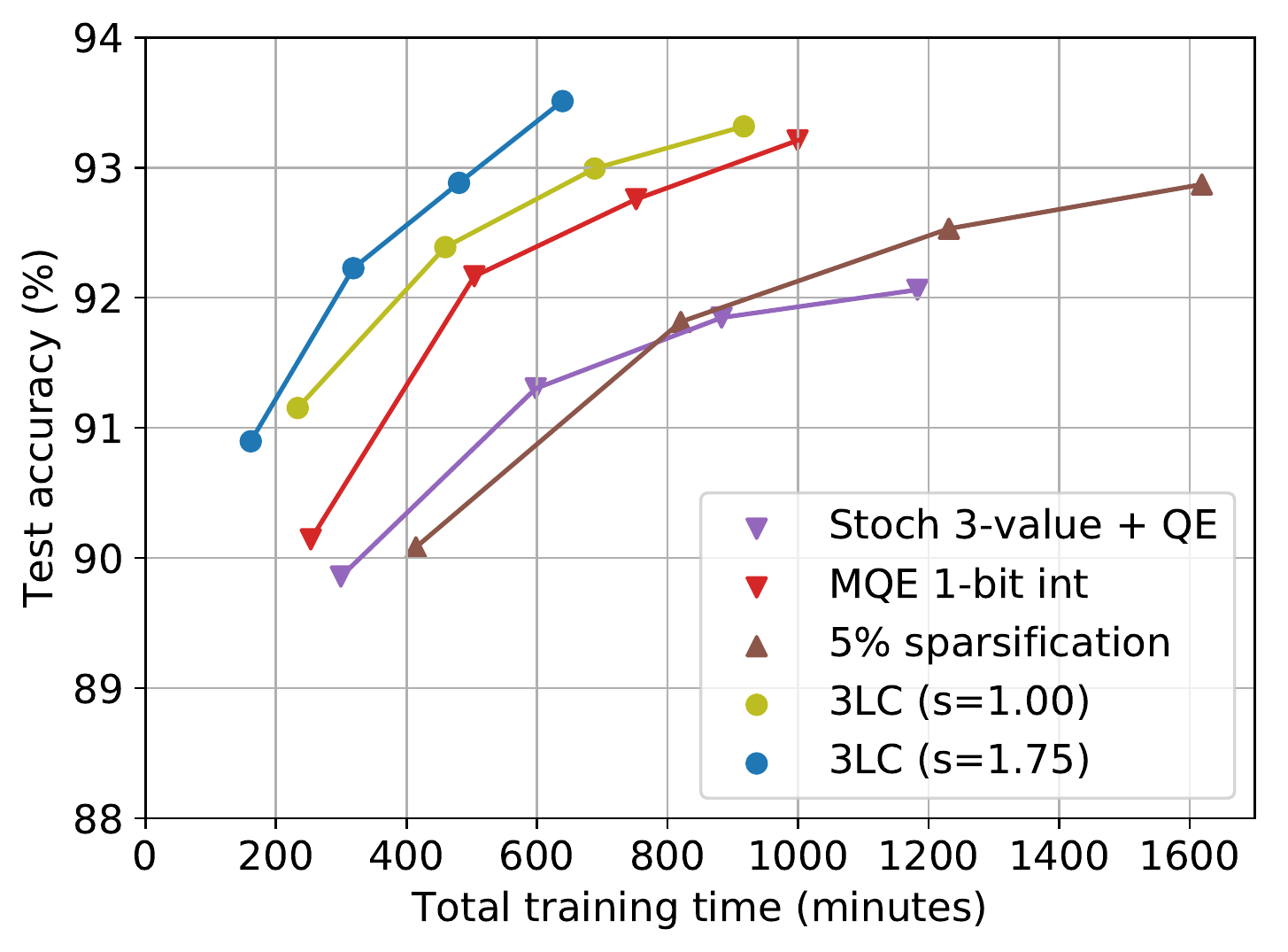}
    \caption{Fast designs}
  \end{subfigure}%
  \caption{Training time and test accuracy using 25/50/75/100\% of standard training steps @ 10 Mbps.}
  \label{fig:tradeoff-10}

  \centering
  \begin{subfigure}[b]{0.5\textwidth}
    \centering
    \includegraphics[width=\textwidth]{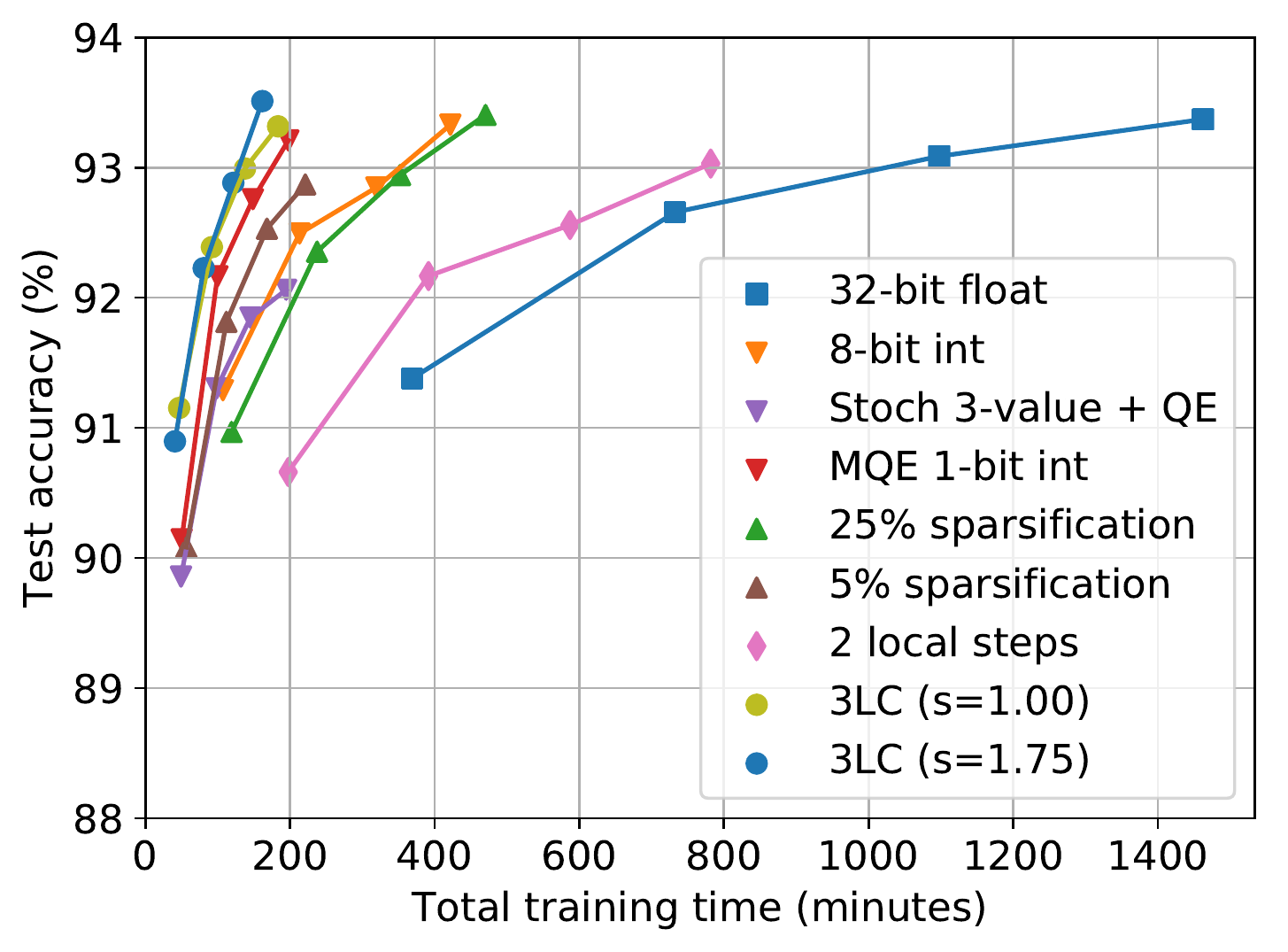}
    \caption{Overview}
  \end{subfigure}%
  ~
  \begin{subfigure}[b]{0.5\textwidth}
    \centering
    \includegraphics[width=\textwidth]{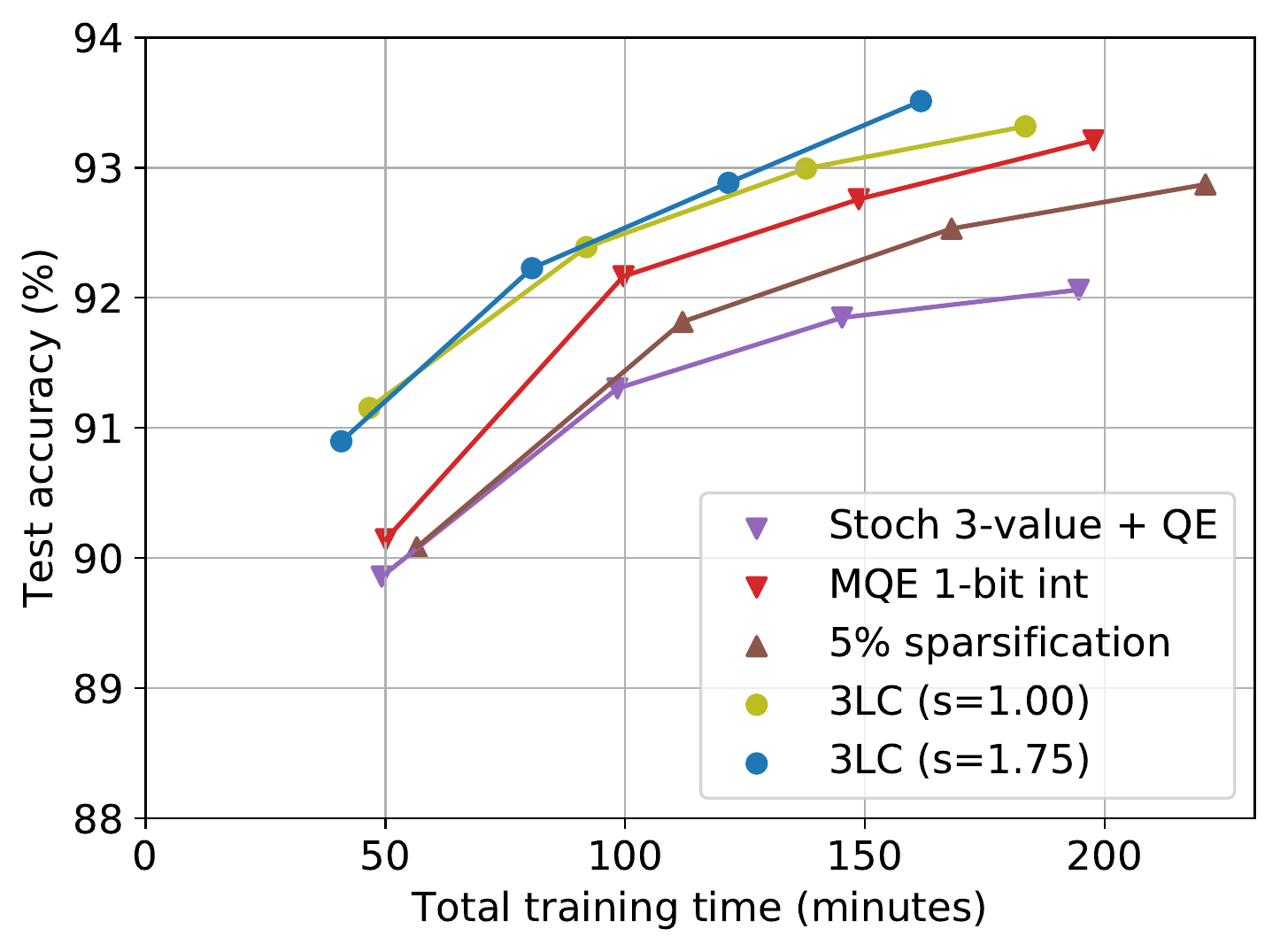}
    \caption{Fast designs}
  \end{subfigure}%
  \caption{Training time and test accuracy using 25/50/75/100\% of standard training steps @ 100 Mbps.}
  \label{fig:tradeoff-100}

  \centering
  \begin{subfigure}[b]{0.5\textwidth}
    \centering
    \includegraphics[width=\textwidth]{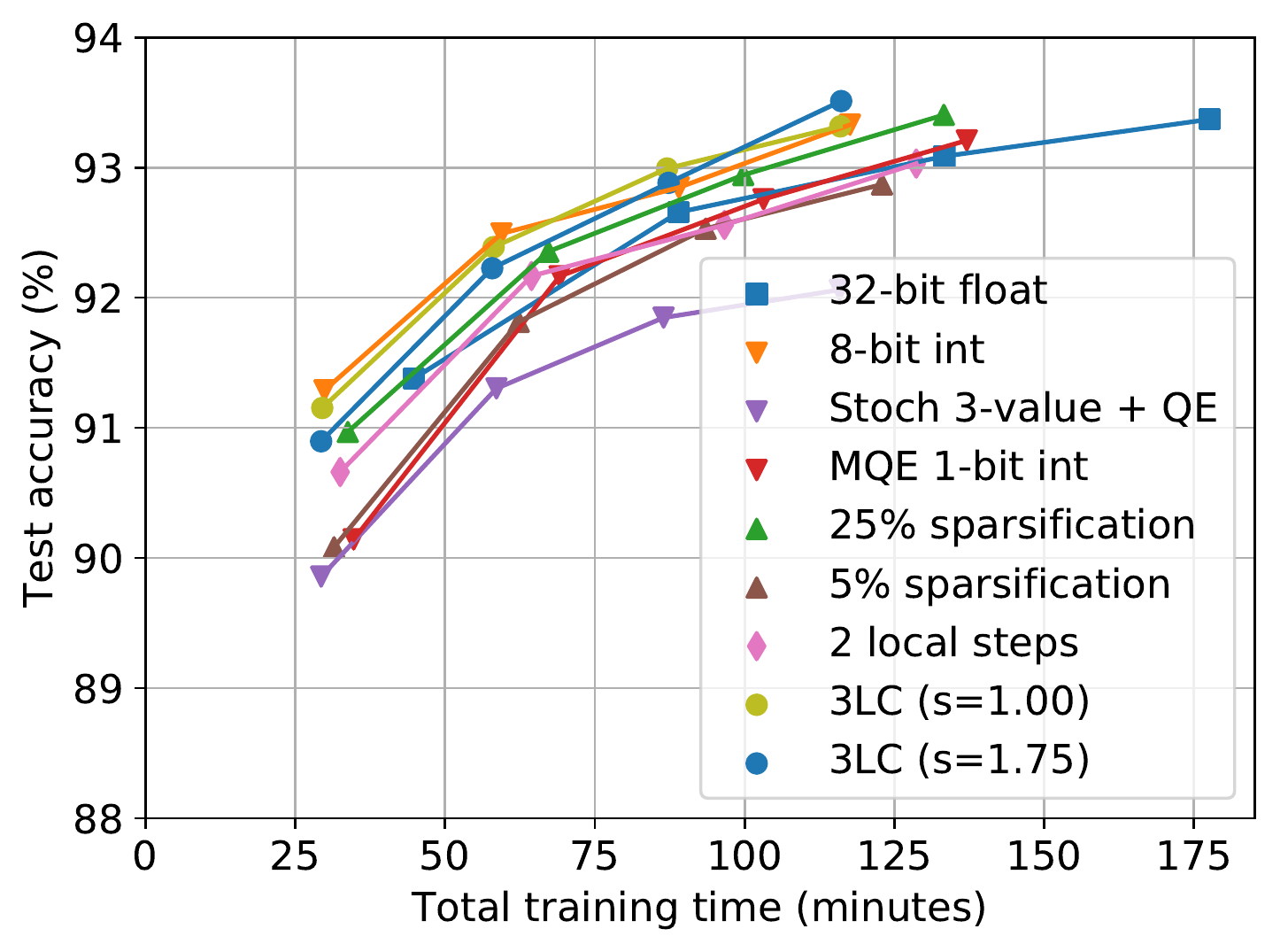}
    \caption{Overview}
  \end{subfigure}%
  ~
  \begin{subfigure}[b]{0.5\textwidth}
    \centering
    \includegraphics[width=\textwidth]{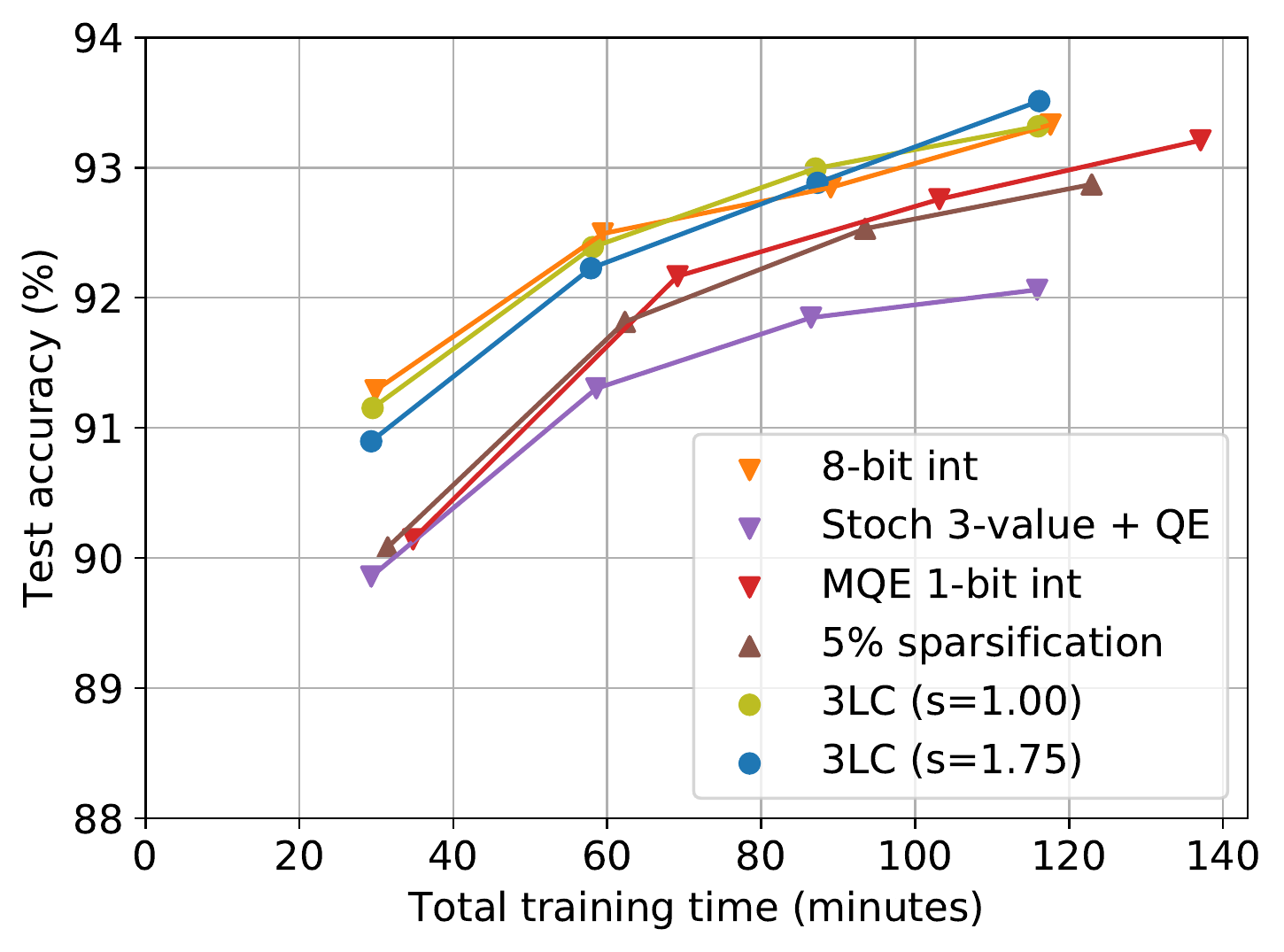}
    \caption{Fast designs}
  \end{subfigure}%
  \caption{Training time and test accuracy using 25/50/75/100\% of standard training steps @ 1 Gbps.}
  \label{fig:tradeoff-1000}
\end{figure*}

\begin{figure*}[t]
  \centering
  \includegraphics[width=0.5\textwidth]{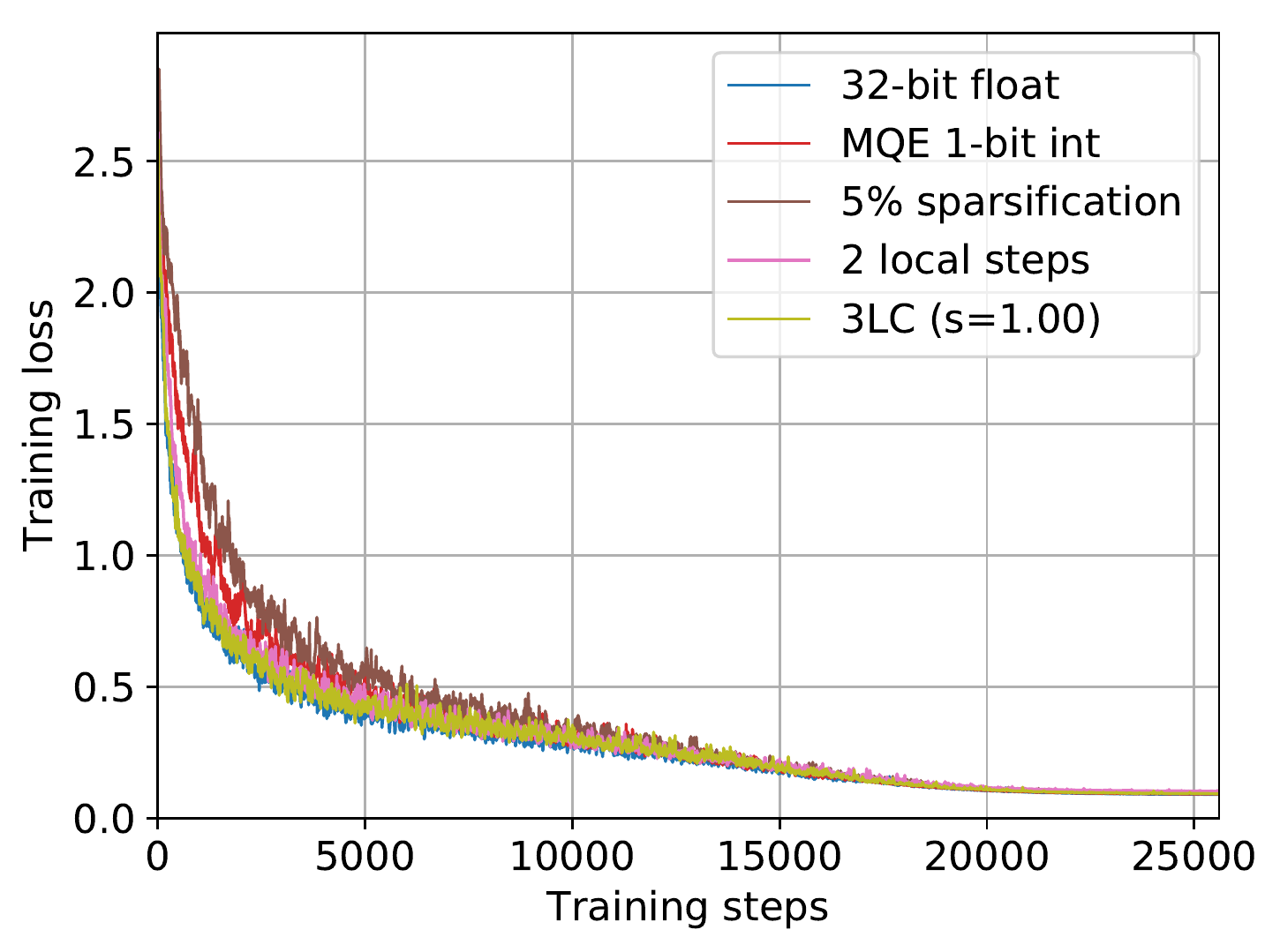}%
  ~
  \includegraphics[width=0.5\textwidth]{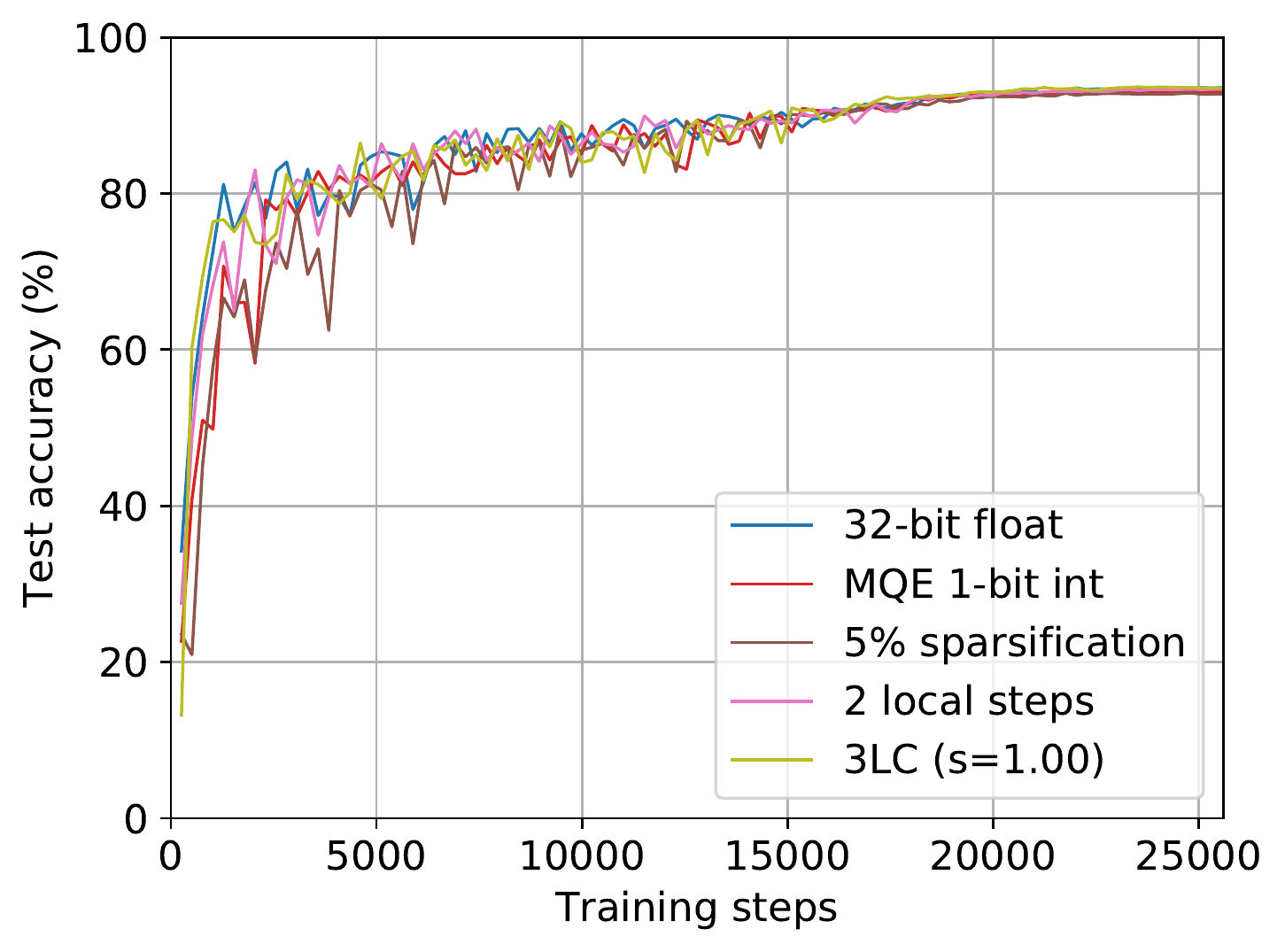}%
  \caption{Training loss (left) and test accuracy (right) using standard training steps.}
  \label{fig:timeline}
\end{figure*}

We also examine how designs perform during a training run in detail.
Figure~\ref{fig:timeline} depicts runtime (not final) training loss and test
accuracy of the baseline, the most representative quantization, sparsification,
and multiple local steps designs, and \name with the default sparsity multiplier;
the result of omitted designs is similar to that of a close design (e.g.,
\texttt{8-bit int} is similar to the baseline).
Except for \name, traffic reduction designs tend to have higher training loss,
and their accuracy also increases slowly.  In contrast, \name achieves small
training loss and high accuracy that are close to those of the baseline.

\begin{figure}[t]
  \centering
  \includegraphics[width=\columnwidth]{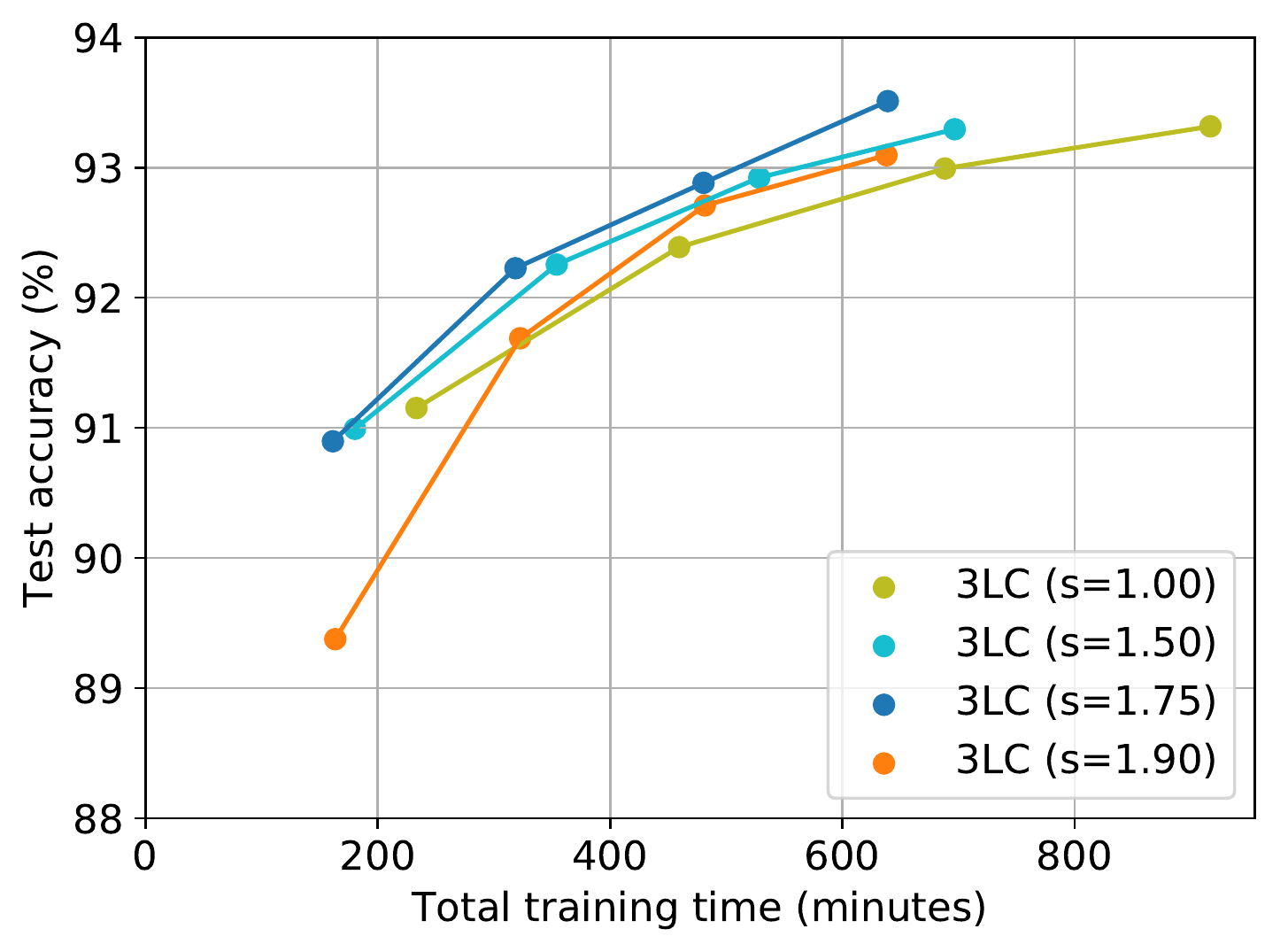}
  \caption{Training time and test accuracy with a varied sparsity multiplier (s) using 25/50/75/100\% of standard training steps @ 10 Mbps.}
  \label{fig:tradeoff-10-sm}
\end{figure}

\begin{figure*}[t]
  \centering
  \includegraphics[width=0.5\textwidth]{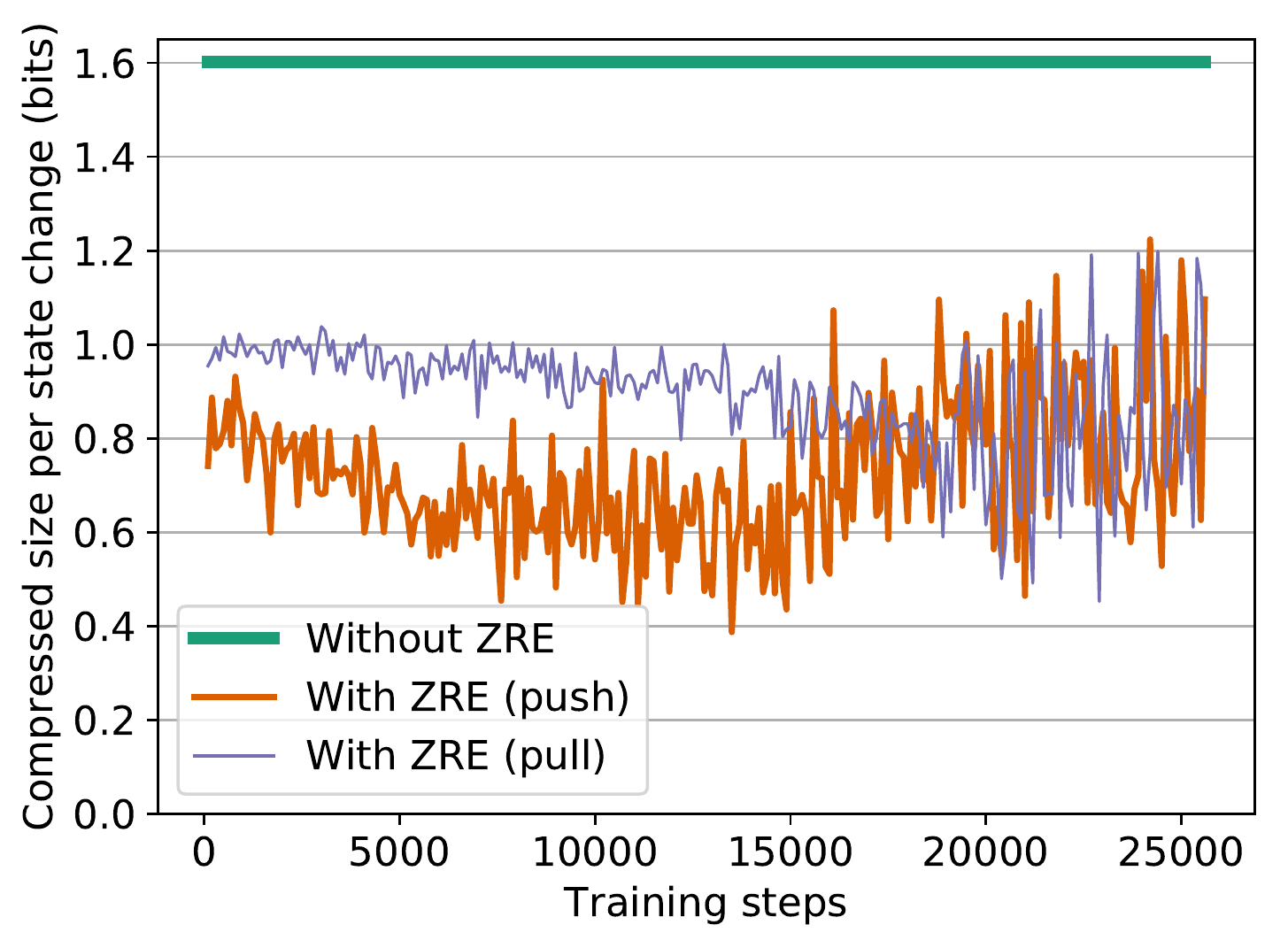}%
  ~
  \includegraphics[width=0.5\textwidth]{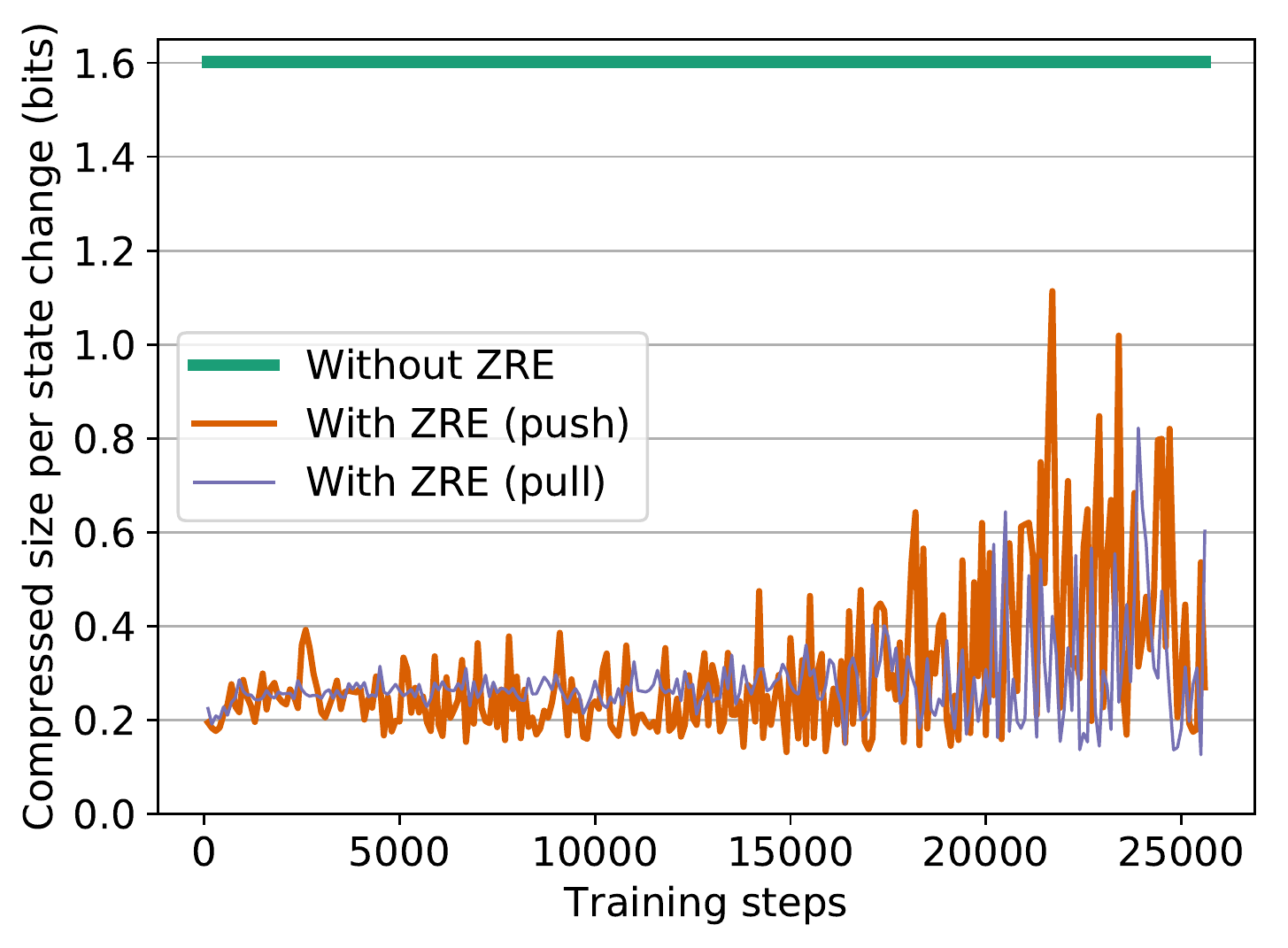}%
  \caption{Compressed size per state change value using standard training steps (left: s=1.00; right: s=1.75).}
  \label{fig:size-sm}
\end{figure*}

\subsection{Sensitivity Analysis}

The control knob of \name is a sparsity multiplier $s$. With a high $s$, 3-value
quantization emits more zeros that can make zero-run encoding more effective.
We vary $s$ and measure training time, traffic reduction, and accuracy.

\begin{table}[t]
  \small
  \centering
  \begin{tabular}{c | S[table-format=3.2] S[table-format=3.2]}
    {$s$}    & {Compression ratio ($\times$)} & {bits per state change} \\ \hline
    {No ZRE} & 20.0                & 1.60               \\
    1.00     & 39.4                & 0.812              \\
    1.50     & 70.9                & 0.451              \\
    1.75     & 107                 & 0.298              \\
    1.90     & 160                 & 0.200              \\
  \end{tabular}
  \caption{Average traffic compression of \name using standard training steps.}
  \label{tbl:tradeoff-sm}
\end{table}

Figure~\ref{fig:tradeoff-10-sm} compares tradeoffs between total training time
and test accuracy.  In general, a high sparsity multiplier reduces training
time, but it can also lower convergence speed with fewer training steps.
Most $s$ values lead to high accuracy when using 100\% of standard training
steps, but $s=1.90$ exhibits lower accuracy than others.

Table~\ref{tbl:tradeoff-sm} examines the average traffic reduction of \name.
Without zero-run encoding (``ZRE''), the quartic-encoded size of each state
change is $1.6$ bits.  Applying zero-run encoding halves the traffic volume for
the default sparsity multiplier ($s=1.00$).  With a higher $s$, \name can
compress traffic more aggressively; \texttt{\name (s=1.90)} realizes a
$160\times$ end-to-end compression ratio and $0.2$ bits per state change.  This
high compression ratio can be useful for metered and/or highly
bandwidth-constrained network connections where reducing the number of bytes
required for state change transmission is crucial for cost-effective
distributed ML\@.

The compression ratio of zero-run encoding changes over time because nodes
generate different gradients and model deltas as the model changes.
Figure~\ref{fig:size-sm} plots the compressed size of gradient pushes and model
pulls at each training step when executing standard training steps.  Compressed
pushes tend to be smaller than compressed pulls until the later stage of
training, which indicates that state changes in model pulls have lower variance
(including fewer zeros in the quantization output) because these changes reflect
aggregated gradient pushes from multiple workers.  After finishing approximately
70\% of training, compressed pushes become larger, which shows that workers now
generate gradients with lower variance.  \name does not forcefully limit how
many state changes can be transmitted at each training step; it permits
transmitting important state changes as much as needed, which can help achieve
fast convergence and high accuracy.
\makeatletter{}%
\makeatletter{}%
\section{Related Work}
\label{sec:related}

\para{Quantization:} 1-bit stochastic gradient
descent~\cite{Seide:interspeech2014} represents each state change with two
values, which can be dequantized using two floating-point numbers that minimize
squared quantization errors.  It accumulates quantization errors for later error
correction.  \name provides more effective traffic reduction that transmits
approximately 1.6-bit worth information in a sub-1-bit representation without
reducing the maximum magnitude of state change values (important for fast
convergence and high accuracy). \name also provides a sparsity multiplier that
can change its compression level.  \name's quantization and encoding methods are
easier to vectorize by using existing vectorizable operations.

QSGD~\cite{Alistarh:nips2017} and TernGrad~\cite{Wen:nips2017} use stochastic
quantization that makes quantized values an unbiased estimator of the original
values.  \name uses error accumulation buffers that empirically provide better
accuracy without introducing changes to machine learning algorithms for accuracy
compensation.

TernGrad~\cite{Wen:nips2017} uses 3-values to quantize state changes, which is
similar to \name's 3-value quantization.  However, TernGrad lacks a knob to
control the compression level and introduces a barrier to synchronize quantization
parameters across all workers.  TernGrad uses 2-bit encoding, which is far less
compact than \name's encoding that requires only 0.3--0.8 bits per state change.

Quantization methods often employ entropy coding schemes such as Huffman coding
and Elias coding for compact binary
representations~\cite{Oland:icassp2015,Alistarh:nips2017}.  \name's zero-run
encoding offers high compression ratios (up to $8\times$) by using byte-level
operations and no lookup tables, which helps achieve low computation overhead.

\para{Sparsification:} The parameter server~\cite{Li:osdi2014} discusses
filtering zero gradients for small-value model parameters.  \name provides compression
for both gradients and model deltas regardless of the magnitude of the model
parameters.

B\"osen~\cite{Wei:socc2015} can prioritize sending large gradients and
model deltas by sorting them.  Because sorting millions of state change values
is expensive, there are proposals that use a relative
threshold~\cite{Hsieh:nsdi2017}, a global threshold~\cite{Aji:emnlp2017},
per-tensor thresholds~\cite{Lin:arxiv2017}, or round-robin
selection~\cite{Watcharapichat:socc2016} for low-overhead sparsification.  Among
these, Gaia~\cite{Hsieh:nsdi2017} changes the relative threshold to send more
state changes as training progresses.  \name transmits larger compressed data in
the later stage of training without having to control the compression level
explicitly.

Gradient dropping~\cite{Aji:emnlp2017} and Deep Gradient
Compression~\cite{Lin:arxiv2017} achieve high compression by selecting only
0.1\% of gradients.  This very aggressive gradient reduction, however, has worse
accuracy.  Recovering accuracy necessitates modifying machine learning
algorithms~\cite{Lin:arxiv2017}, which reduces their generality and makes it
hard to compress non-gradient state changes such as model deltas.

Project ADAM~\cite{Chilimbi:osdi2014} and Poseidon~\cite{Zhang:atc2017} reduce
network traffic by transmitting small ``sufficient factors'' that contain enough
information to construct full gradient tensors for certain types of neural
network layers~\cite{Xie:arxiv2014}.  \name pursues a general tensor compression
scheme that can compress gradients and model deltas for any type of layers.

\para{Infrequent communication:} Federated
learning~\cite{Konevcny:arxiv2016,McMahan:aistats2017} runs multiple training
steps before each global state change transmission.  Our evaluation shows that
infrequent transmission of state changes can lead to lower accuracy when using
the same number of training steps.

\makeatletter{}%
\section{Conclusion}
\label{sec:concl}

A key challenge in modern, large-scale machine learning is marrying the demands
of systems (reducing communication, overlapping computation and communication,
and so on) and learning algorithms (algorithmic efficiency, accuracy, and
convergence).  In this paper, we described a new lossy compression scheme for
distributed training of machine learning models that reduces network traffic by
up to $107\times$ without impairing training or altering machine learning
algorithms.  The key contribution is a new traffic compression scheme that
combines the strengths of tensor quantization and sparsification approaches.
\name introduces three new lightweight, yet effective lossy and lossless
transformation techniques, resulting in greater balance between traffic
compression, accuracy, computation overhead, and generality in distributed
training under a large range of available network bandwidths.

\clearpage

\setlength{\bibsep}{2pt plus 1pt}  %
\small
\balance
}{%
}

\end{document}